\documentclass{ieeeaccess}
\usepackage{cite}
\usepackage{amsmath,amssymb,amsfonts}
\usepackage{algorithmic}
\usepackage{graphicx,booktabs}
\usepackage{textcomp}
\usepackage{bm}
\usepackage{multirow}

\ifCLASSOPTIONcompsoc
\usepackage[caption=false, font=normalsize, labelfont=sf, textfont=sf]{subfig}
\else
\usepackage[caption=false, font=footnotesize]{subfig}

\def\BibTeX{{\rm B\kern-.05em{\sc i\kern-.025em b}\kern-.08em
    T\kern-.1667em\lower.7ex\hbox{E}\kern-.125emX}}
\begin{document}
\history{Date of publication xxxx 00, 0000, date of current version xxxx 00, 0000.}
\doi{xxxx/ACCESS.xxxx}

\title{Triplet Online Instance Matching Loss for Person Re-identification}
\author{\uppercase{Ye Li}\authorrefmark{1},
\uppercase{Guangqiang Yin\authorrefmark{1}, Chunhui Liu\authorrefmark{1}, Xiaoyu Yang\authorrefmark{1}, and Zhiguo Wang\authorrefmark{1}}}

\address[1]{University of Electronic Science and Technology of China. (e-mail: liye@std.uestc.edu.cn, yingq@uestc.edu.cn, lettherebe741@std.uestc.edu.cn, yangxy@std.uestc.edu.cn, zgwang@uestc.edu.cn)}

\markboth
{LiYe \headeretal: Triplet Online Instance Matching Loss for Person Re-identification}
{LiYe \headeretal: Triplet Online Instance Matching Loss for Person Re-identification}

\corresp{Corresponding author: GUANGQIANG YIN (e-mail: yingq@uestc.edu.cn).}

\begin{abstract}
Mining the shared features of same identity in different scene, and the unique features of different identity in same scene, are most significant challenges in the field of person re-identification (ReID). Online Instance Matching (OIM) loss function and Triplet loss function are main methods for person ReID. Unfortunately, both of them have drawbacks. OIM loss treats all samples equally and puts no emphasis on hard samples. Triplet loss processes batch construction in a complicated and fussy way and converges slowly. For these problems, we propose a Triplet Online Instance Matching (TOIM) loss function, which lays emphasis on the hard samples and improves the accuracy of person ReID effectively. It combines the advantages of OIM loss and Triplet loss and simplifies the process of batch construction, which leads to a more rapid convergence. It can be trained on-line when handle the joint detection and identification task. To validate our loss function, we collect and annotate a large-scale benchmark dataset (UESTC-PR) based on images taken from surveillance cameras, which contains 499 identities and 60,437 images. We evaluated our proposed loss function on Duke, Marker-1501 and UESTC-PR using ResNet-50, and the result shows that our proposed loss function outperforms the baseline methods by a maximum of 21.7\%, including Softmax loss, OIM loss and Triplet loss.
\end{abstract}


\titlepgskip=-15pt

\maketitle

\section{Introduction}
\PARstart{P}{erson} re-identification (ReID) is mainly used for distinguishing identity of a pedestrian on different cameras. It focuses more on the whole image of pedestrians than face recognition. It is widely used in public security, including intelligent security, intelligent video surveillance and criminal investigation. However, it is quite difficult to distinguish identities as lighting, occlusion, camera viewpoints and background clutter vary greatly with different cameras and conditions, especially the examples with same identity but extremely different features or the ones with different identities but extremely similar features. These examples are named as "Hard samples" (see Fig.~\ref{hard_sample}) and has arouse a new research problem about the method to tackle them effectively.

\Figure[!htpb](topskip=0pt, botskip=0pt, midskip=0pt)[width=3.0in]{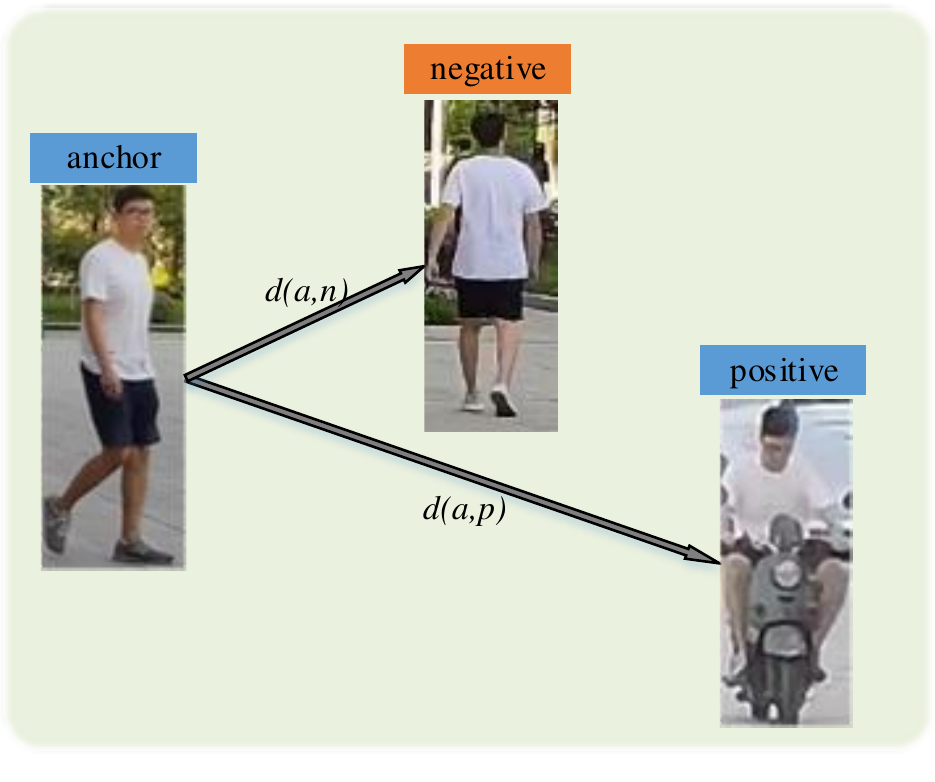}
{A set of "Hard example" : anchor and positive are the same ID, and the distance between the features is denoted by $d\left ( a,p \right )$; anchor and negative are different ids, and the distance between features is denoted by $d\left ( a,n \right )$. But $d\left ( a,p \right )\gg d\left ( a,n \right )$.\label{hard_sample}}
\begin{figure*}[!htpb]
    \centering
	  \subfloat[Feature distribution using TOIM]{
       \includegraphics[width=0.3\linewidth]{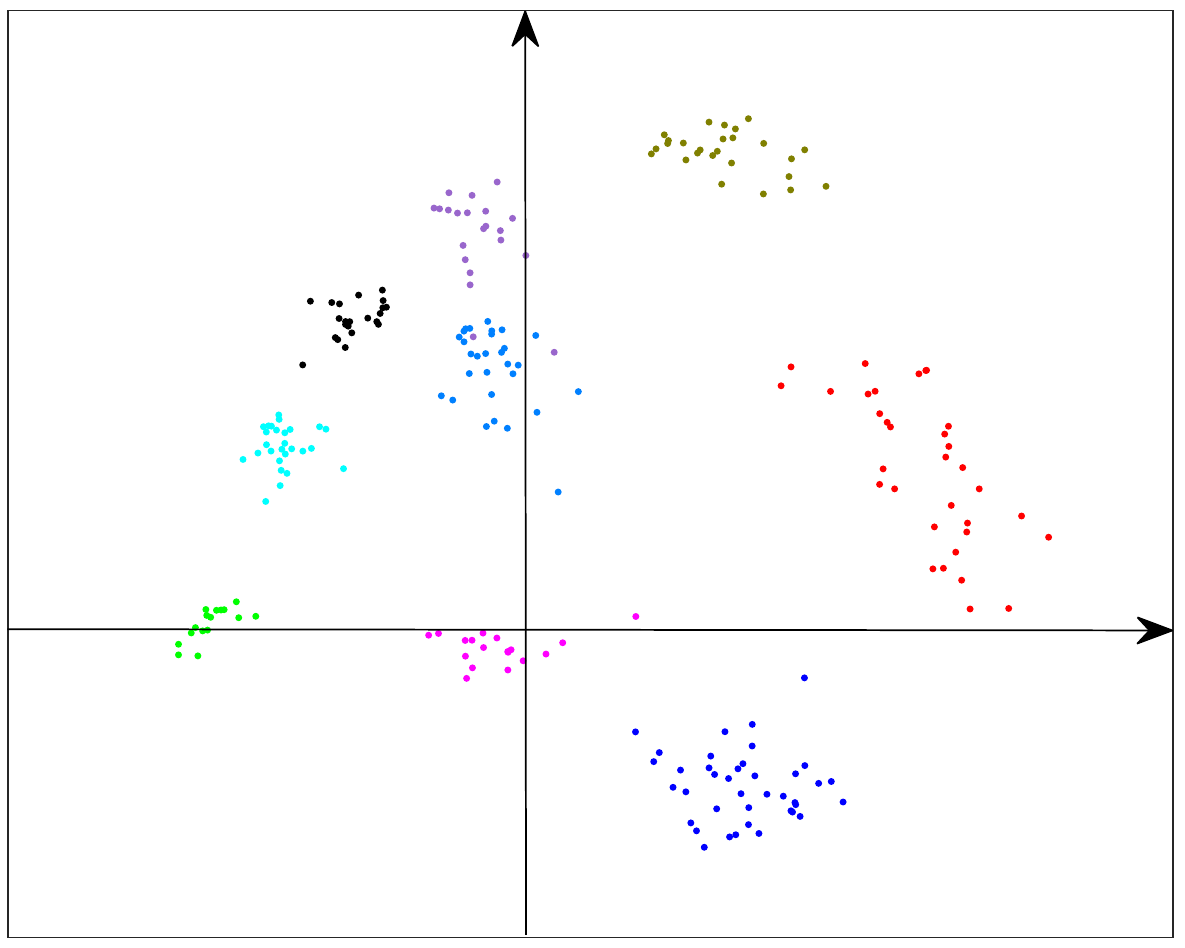}}
    \label{}\hfill
	  \subfloat[Feature distribution using Softmax]{
        \includegraphics[width=0.3\linewidth]{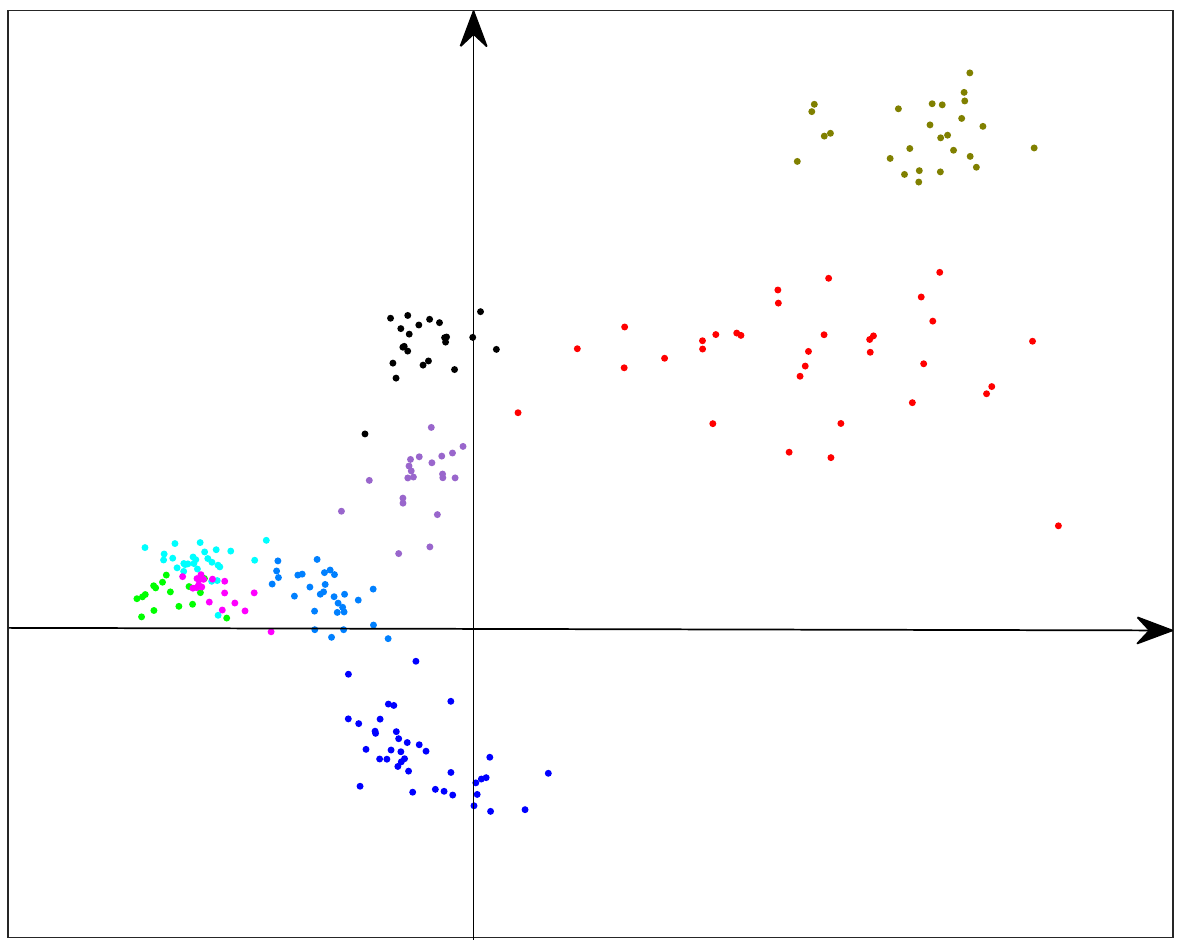}}
    \label{}\hfill
	  \subfloat[Feature distribution using OIM]{
        \includegraphics[width=0.3\linewidth]{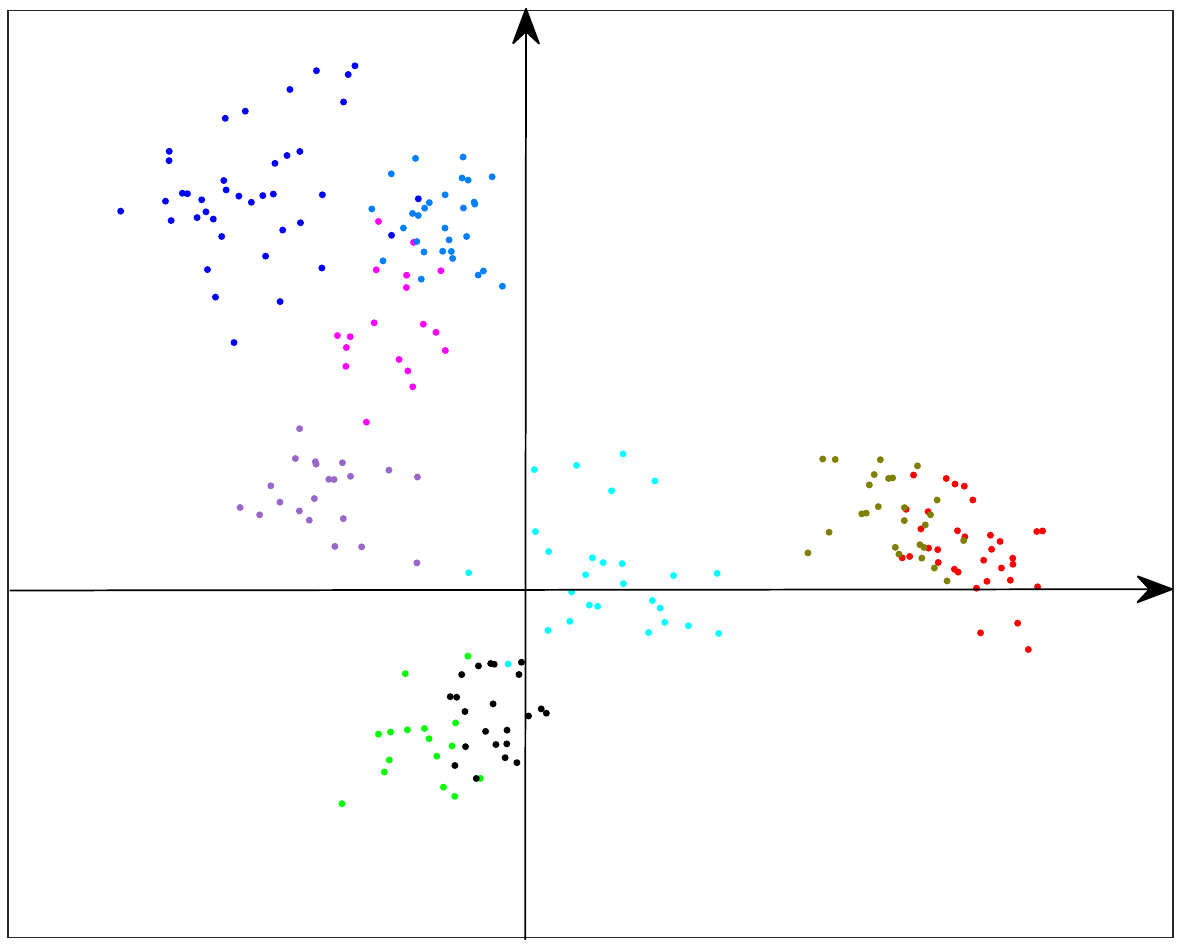}}
    \label{}\hfill
	  \caption{(a), (b), (c) Feature distribution in two dimensions. Different colors represent different identities.}
	  \label{point_distribution}
\end{figure*}

The general concept of person re-identification can be divide into three modules: person detection, person tracking and person retrieval \cite{Zheng2016}. For one thing, the poor performance of the person detection and person tracking algorithm may impose a negative effect on person retrieval, thus leading to a low accuracy. For another thing, most of the existing benchmarks only contains manually cropped pedestrian images. As a result, many scholars are devoted to the study of person retrieval module itself, which can eliminate the interference of previous two modules and focus on polishing algorithms. In this paper, if not specified, person re-identification (ReID) refers to the person retrieval.

The process of person ReID generally involves three important parts: feature extraction, feature aggregation and loss function. Existing person ReID methods are mainly based on Softmax loss function, Online Instance Matching (OIM) loss function, Triplet loss function, etc. \cite{Wu2019}. Hermants \textit{et al}. \cite{Cheng2016} proposed Triplet loss function, which could conduct hard sample mining in each epoch. It proves that Triplet loss is superior to other loss function in terms of ReID. Xiao \textit{et al}. \cite{Xiao2017} proposed Online Instance Matching (OIM) loss and proved its superiority to Softmax loss. By saving and updating the features of each identity during the training process, it can be trained on-line when handling the joint detection and identification task. However, there exist some certain limitations in both Triplet loss and OIM loss. As for the Triplet loss, firstly, its process of constructing batch is relatively cumbersome, as each batch must contain P identities, and for each identity it must contain K samples; Secondly, it usually takes a longer time to converge; Thirdly, it cannot handle the joint detection and identification task when trained on-line. The limitation of OIM is that all person samples are treated equally, and it puts no emphasis on hard samples.

To solve the problems above, we propose a new loss function called Triplet Online Instance Matching (TOIM), which combines the advantages of Triplet loss and OIM loss. In this paper, pedestrian samples are first sent into the feature extraction network. Then, the extracted features are stored in Pooled Table for further use. Finally, TOIM loss function is used to train the network. TOIM loss can optimize the embedding space such that data points with the same identity are closer to each other than those with different identities, and the distribution of most samples in the embedding space is further away from zero point.

To verify the robustness of the TOIM loss function, a new dataset UESTC-PR is made based on the real surveillance images (CAM overlooking the target). In order to evaluate the performance of TOIM compared with other loss functions, we adopt ResNet-50 \cite{He2016} as the basic network to extract pedestrian features. Four loss functions are used to train the network respectively, namely OIM, Softmax, Triplet and TOIM. The experiment shows that TOIM loss function can significantly improve the accuracy of pedestrian re-identification, and it shows a noticeably better performance than other loss functions. To visualize the effect of TOIM, we randomly selected several samples and then reduced their features to two dimensions using Principal Component Analysis (PCA). The two-dimensional distributions of features obtained by TOIM, Softmax and OIM loss are shown in Fig.~\ref{point_distribution}.

The major contributions of this paper are summarized as follows:
\begin{itemize}
  \item We propose a new loss function (TOIM) to learn identification features more effectively. We validate the effectiveness of our loss function comparing with other loss function;
  \item We collect and annotate a large-scale benchmark dataset for person re-identification, and demonstrate that TOIM loss is very robust to the noise in a real surveillance video or image.
\end{itemize}

The rest of the paper is organized as follows. Two related work, OIM loss and Triplet loss, are introduced in Section \ref{sec:2}. In Section \ref{sec:3}, we propose our method, TOIM loss, and describe it in details. The datasets invoked in this paper are presented in Section \ref{sec:4}. We implement experiments to prove the advantages of our proposed method in Section \ref{sec:5}. And a conclusion is drawn in Section \ref{sec:6}. 

\section{Related work}\label{sec:2}
\subsection{OIM for Person Re-identification}
OIM loss function, which is widely used in joint detection and identification task, was proposed together with an end-to end ReID model based on Faster-CNN by Xiao \cite{Xiao2017}. According to different type of proposals, training samples are divided into different groups: labeled identities, unlabeled identities and background clutter. The labeled identities are stored in LookUp Table (LUT) and continuously updated by the specific renewal strategy, while unlabeled identities are stored in Circular Queue (CQ) and updated by queue. Based on these two structures, OIM loss is defined.

OIM loss is similar to cross-entropy loss. However, there are generally more than 5,000 identities in ReID task and Softmax can only deal with around 10 identities in each iteration, it is prone to suffer from large variance of gradients and fail to learn effectively. In addition, the unlabeled identities cannot be exploited with Softmax due to the deficiency of corresponding class-ids. For the problems above, Xiao \textit{et al}. \cite{Xiao2017} made a leap forward by modifying Softmax loss into non-parametric. The proposed LUT and circular queue serve as buffers from the outside rather than inner parameters, thus enabling the gradients to act on features directly. Several comparison experiments have shown its superiority over Softmax.

\subsection{Triplet Loss for Person Re-identification}
Triplet loss was originally proposed in \cite{Schroff2015}, which was initially aimed at face recognition and clustering by exploited triplet samples for training network to make similar faces cluster together. Triplet loss is widely used in ReID for its specific function of enabling data points with the same identity to become closer than those with different identities. Hermans \textit{et al}. \cite{Cheng2016} presented a variant of the Triplet loss, which outperforms most other state-of-art methods by a large margin. Wang \textit{et al}. \cite{Wang2016} proposed a new learning framework that combined Single-Image Representation (SIR) and Cross-Image Representation (CIR). However, it performed no function of hard sample mining and pays equal attentions to all the triplets. Cheng \textit{et al}. \cite{Cheng2016} proposed a convolutional neural network that could jointly learn the features of global full-body and local body-parts. Li \textit{et al}. \cite{Li2019} proposed a spatiotemporal feature extraction method, which inserted Non-local blocks into the 3D residual network, the loss function of which consists of a cross-entropy loss and a Triplet loss. \cite{Wu2019-Omni} Proposed that model can learn part representation with spatial information from vertical and horizontal orientations. And the triplet loss and OIM loss are employer to learn a similarity measurement and global features at the same time.

\section{Method}\label{sec:3}
\subsection{model structure}
In this paper, we adopt ResNet-50 as basic network which consists of one $7\times 7$ convolution layer (Conv\_1) and four modules, namely \textit{res}$_2$, \textit{res}$_3$, \textit{res}$_4$ and \textit{res}$_5$. Each module contains 3, 4, 6 and 3 residual blocks. At the end of the last module, an average pooling is performed and then a fully connected layer is attached. The exact network configuration is shown in Table 1. The input image has a dimension of $N\times 3\times 256\times 128$, where $N$ is batchsize, 3 is the number of channels, $256\times 128$ is the size of image. The output result is $N\times 512$, where 512 represents number of features in each image and all these 512 features are stored respectively in Pooled Table.


\Figure[!htpb](topskip=0pt, botskip=0pt, midskip=0pt)[width=7in]{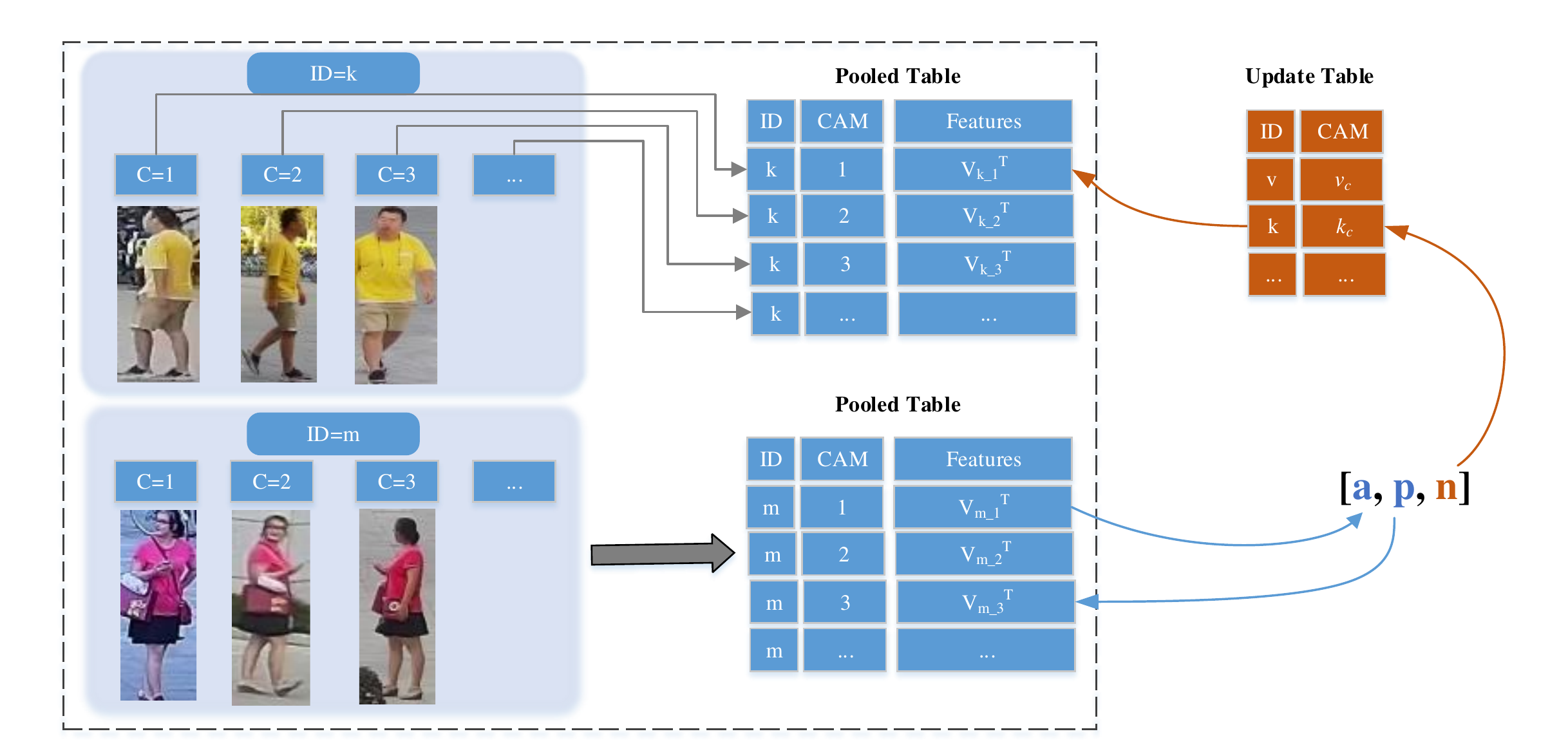}
{TOIM Loss. \textbf{Left}: the Pooled Table for storing the feature of different ID and different CAM. \textbf{Right}: selecting features from Update Table to calculate loss.\label{TOIM}}


\begin{table}[!htbp]
        \centering
         \caption{Our ResNet-50 model for ReID. The size of the input image is $256\times128$.}
         \setlength{\tabcolsep}{3.5mm}{
        \begin{tabular}{cccc}
         &layer	& &output size \\
         \toprule
         \textit{conv}$_1$&	$7\times7, 64$, stride $2$, $2$&$\times1$&$128\times64$  \\
         \midrule
         \textit{pool}$_1$&	$3\times3$ avg, stride $2$, $2$&$\times1$&$64\times32$ \\
         \midrule
         &$1\times1$, 64& & \\
         \textit{res}$_2$&$3\times3$, 64 & $\times3$&	$64\times32$ \\
         &$1\times1$, 256& &\\
         \midrule
         &$1\times1$, 128& & \\
         \textit{res}$_3$&$3\times3$, 128 & $\times4$&	$64\times32$ \\
         &$1\times1$, 512& &\\
         \midrule
         &$1\times1$, 256& & \\
         \textit{res}$_4$&$3\times3$, 256 & $\times6$&	$64\times32$ \\
         &$1\times1$, 1024& &\\
         \midrule
         &$1\times1$, 512& & \\
         \textit{res}$_5$&$3\times3$, 512 & $\times3$&	$64\times32$ \\
         &$1\times1$, 2048& &\\
         \midrule
         &	global average pool, fc &$\times1$ &$1\times1\times1$ \\
         \bottomrule
       \end{tabular}}
\end{table}

\subsection{Triplet Online Instance Matching Loss (TOIM)}\label{sec:3.2}
The aim of ReID is to distinguish pedestrians of different identities. One should clustering the features according to their identities by minimizing the distances between same identity and increasing the distances between different identities. Though OIM loss can achieve this purpose to a certain extent, all samples are treated equally and no emphasis is laid on samples that are extremely difficult to handle, for OIM loss is proposed on the basis of a cross-entropy loss function. Though Triplet loss has specific training process targeting at the hard samples, it will not continue to increase once the margin reaches the critical value. As a Hinge loss function, Triplet is not in line with the purpose of narrowing the intra-class distance and increasing the inter-class distance in ReID’s task. Therefore, by combining the idea of cross entropy loss and advantages of Triplet loss, we propose the TOIM loss function. It not only enables the target to be continuously optimized, but also distinguish the hard samples with a high accuracy.

To enable TOIM to select specific samples for training, we first adopt Pooled Table (PT) to store the features of pedestrians with different identities and CAMs. The features of one identity is represented as $f\in R^{D}$, where $D$ is the feature dimension. Each identity will generate a Pooled Table $V\in R^{D\times C}$, where $C$ is the number of cameras used in the collection of dataset. If the training set contains $M$ identities, then $M$ Pooled Tables will be generated (see Fig.~\ref{TOIM}). Then, according to the distance between samples, hard samples are selected and combined into triplets ([\textit{a}, \textit{p}, \textit{n}]) for training. Here, \textit{a} stands for anchor; \textit{p} stands for positive, which has the same identity with the anchor; \textit{n} stands for negative, which has a different identity from the anchor. In the process of selecting negative samples, we set the Update Table (UT) to ensure that the loss function can always train on the latest samples. It is a form of queue with the length of $U$ dimensions that can store the ID and CAM of the updated samples in the last batch (see Fig.~\ref{TOIM}). The training process is described in detail below:

During the forward propagation, we randomly select $N$ anchors of different identities. Correspondingly, positive and negative samples are selected for each anchor. For example, for anchor $f_{a}^{i}$ with class-id $i$, we select the sample with the same identity and the largest distance as its positive sample, denoted as $p$ ($f_{p}^{i}$). The sample with different ID and the smallest distance from the Update Table, denoted as $n$ ($f_{n}^{j}$). Then we defined the TOIM loss function as follows:

$$L_{\textit{TOIM}}=\sum_{i=1}^{N}\left ( -\underset{j\neq i}{\log}(\frac{e^{d(f_{a}^{i},f_{n}^{j})}}{e^{d(f_{a}^{i},f_{n}^{j})}+e^{d(f_{a}^{i},f_{p}^{i})}}) \right)  \eqno{(1)}  $$

\noindent where $d()$ is the Euclidean distance between two samples.

During backward propagation, if the target class-id is $p^{*}$ and cam-id is $c^{*}$, we will update the Pooled Table of the $p^{*}$ id and the $c^{*}$ cam by $v_{p^{*}-c^{*}}=\gamma v_{p^{*}-c^{*}}+(1-\gamma)f$, where $\gamma \in [0,1]$.

According to the equation above, we use the selection strategy which we designed to skillfully select three specific samples [\textit{a}, \textit{p}, \textit{n}]. Then we calculate the distance $d(\textit{a},\textit{n})$ between \textit{a} and \textit{n}, the distance $d(\textit{a},\textit{p})$ between \textit{a} and \textit{p}. The idea of cross-entropy loss function to continuously maximize $d(\textit{a},\textit{n})$ and minimize $d(\textit{a},\textit{p})$ is adopted during the training process.
Our TOIM loss effectively distinguishes hard examples, driving the underlying feature to distribute densely within the same identity, while pushing it away from the different identities.

In addition, by saving and updating the features of each identity during training process, TOIM loss function can be trained on-line when handling the joint detection and identification task. During the batch construction of Triplet loss function, each batch must contain P identities and K samples for each identity. However, only N anchors of different identities should be selected randomly in batch construction of TOIM loss function. Then the proposed algorithm selects the corresponding positive and negative sample for each anchor automatically. Thus make the process convenient and unrestricted.	 For the joint detection and identification task, the input of model can only be the single frame which contains different identities. TOIM loss can easily construct batch while Triplet loss fails in the case above.

\section{Dataset}\label{sec:4}
In order to verify the effectiveness of our method, we train our model on two types of datasets: the public dataset DUKE \cite{Zheng2017} and Market-1501 \cite{Zheng2015}, and the UESTC-PR dataset collected by us. In this section, we describe both datasets in detail, and give a brief introduction of ReID's evaluation protocols.

\Figure[!htpb](topskip=0pt, botskip=0pt, midskip=0pt)[width=3.3in]{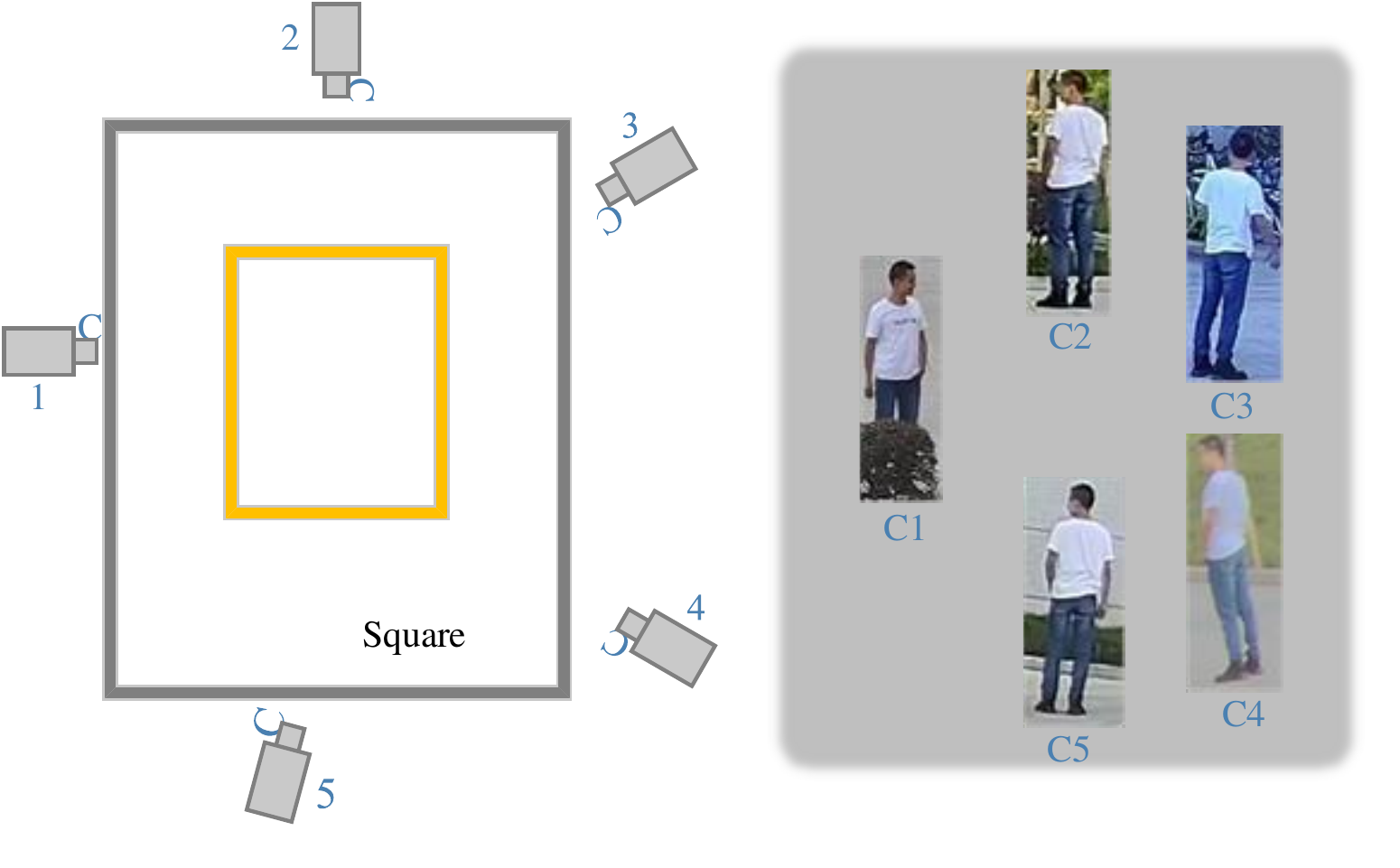}
{Dataset production method and sample examples. \textbf{Left}: the acquisition process of the UESTC-PR dataset. \textbf{Right}: images of the same person under 5 different cameras.\label{uestc-pr}}

\Figure[!htpb](topskip=0pt, botskip=0pt, midskip=0pt)[width=3.3in]{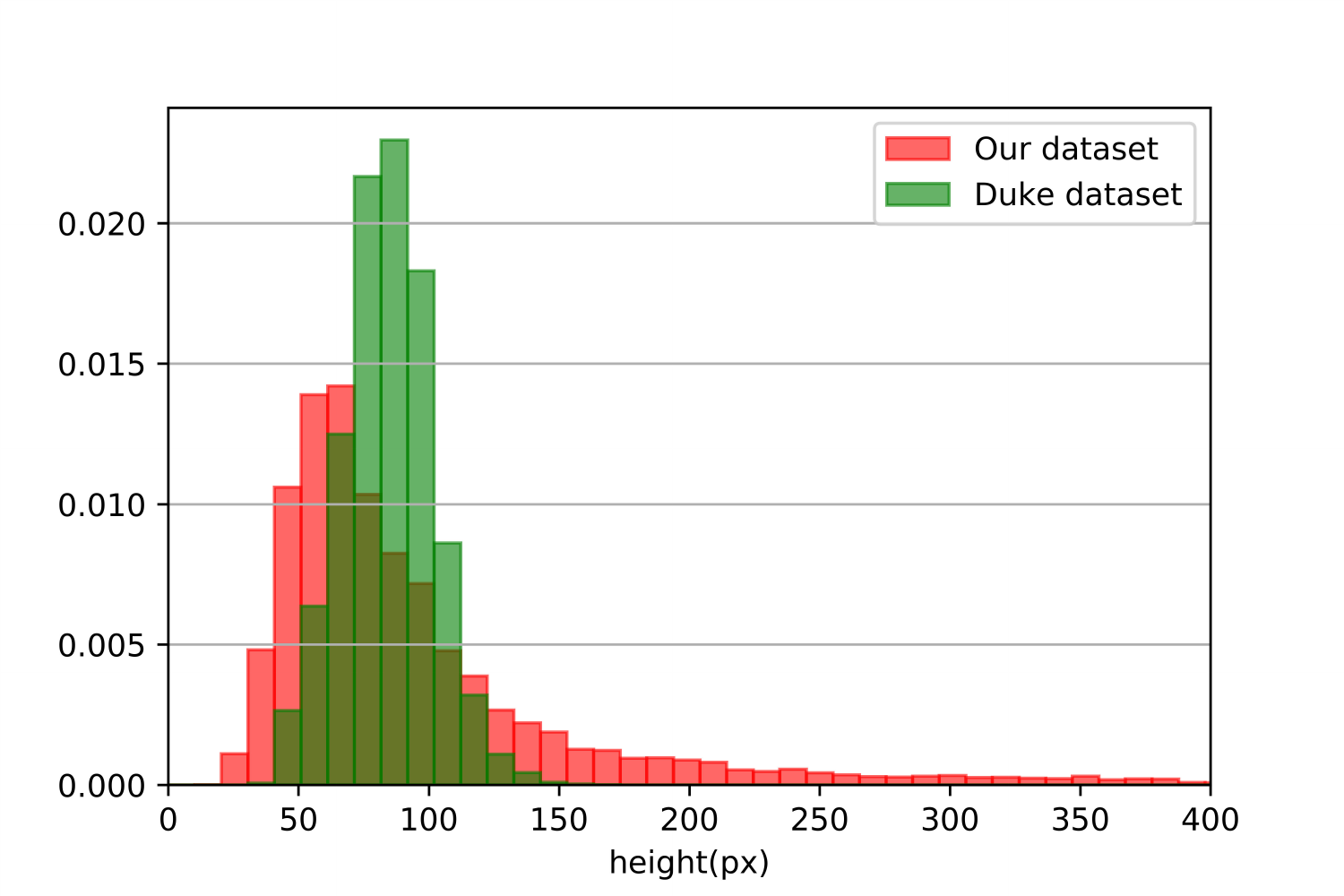}
{The height distributions of Our and Duke datasets.\label{dataset_distribution}}
\label{distributions}

\begin{table}
        \centering
         \caption{Statistics of the dataset of experiments.}
         \setlength{\tabcolsep}{0.6mm}{
       \begin{tabular}{ccccc}
         \toprule
         Dataset & Market-1501 & Duke &  \textbf{UESTC-PR} \\
         \midrule
         Pedestrians & 38639 & 32668 & \textbf{60437}  \\
         Identities & 1812 & 1501 & \textbf{499} \\
         Training ID & 702 & 751 & \textbf{251} \\
         Query ID & 702 & 750 & \textbf{247}\\
         Gallery ID & 1110 & 750 & \textbf{248}\\
         Size & big & big & \textbf{small}\\
         Shooting Angle & Smooth Inspect & Smooth Inspect & \textbf{Overlook}\\
         Motion State & Walking & Walking & \textbf{Walking+Cycling}\\
         Detection + Identification & No & No & \textbf{Yes}\\
         \bottomrule
       \end{tabular}}
\end{table}

\subsection{Duke and Market-1501 dataset}

The Duke dataset is collected by 8 different cameras. The whole dataset contains 36,411 images, including 16,522 for train and 19,889 for test. There are 1,404 persons who show up under more than 2 cameras and we randomly sampled 702 as training set, 702 as test set. The Market1501 dataset is collected by 6 different cameras. The whole dataset contains 32,668 images, including 12,936 for train and 19,732 for test. There are 1,501 IDs and we randomly selected 751 as training set, 750 as test set.

\begin{table*}[!htbp]
\centering
\caption{Comparisons between our loss function and others.}
\setlength{\tabcolsep}{1.1mm}{
\begin{tabular}{cccccccccccc}
\toprule
\multicolumn{1}{c}{ \multirow{2}*{\textit{loss}} }&\multicolumn{3}{c}{Duke}& &\multicolumn{3}{c}{Market-1501}& &\multicolumn{3}{c}{UESTC-PR}\\
\cmidrule{2-4}
\cmidrule{6-8}
\cmidrule{10-12}
\multicolumn{1}{c}{}&mAP&CMC CUHK03&CMC Market-1501& &mAP&CMC CUHK03&CMC Market-1501& &mAP&CMC CUHK03&CMC Market-1501\\
\midrule
Softmax&40.7&44.3&62.5& &62.2&68.9&79.5& &49.3&59.5&72.0\\
OIM&47.4&50.4&68.1& &60.3&68.1&77.1& &53.7&63.9&71.1\\
Triplet&54.6&57.5&73.1& &67.9&70.5&81.1& &54.1&65.8&70.2\\
TOIM&\textbf{62.4}&\textbf{64.2}&\textbf{78.4}& &\textbf{69.2}&\textbf{74.3}&\textbf{83.3}& &\textbf{62.3}&\textbf{71.7}&\textbf{77.7}\\
\bottomrule
\end{tabular}}
\end{table*}

\subsection{UESTC-PR dataset}

Dataset is a very important part in the field of person re-recognition. However, most of the existing datasets only contain head-up targets, which is different from the real-world scenarios. In order to simulate of the actual application scenario and minimize the production cost of the dataset, we proposed a new dataset collection method and collected the UESTC-PR dataset in this way.

The dataset in this paper was collected in the small square of innovation center in University of Electronic Science and Technology of China (UESTC). Five cameras were adopted and distributed in all directions of the central square. The detailed specific planning is shown in Fig. 4. When a pedestrian passes by, different cameras take pictures of different poses and backgrounds of the same identity. Thus it can meet the requirement of person re-recognition for diversified pose of the sample in the dataset. At the same time, due to the large square and the high position of the camera, the acquired pedestrians are small, which is in line with the actual scene. In order to control the impact of light on the accuracy of person recognition, video collection was conducted in three periods of time in a day, namely 8:30-9:30 am, 11:00-12:00 am and 17:00-18:00 pm respectively. For each camera, the recording lasts for 67 minutes and one frame in a second. The dataset covers diverse backgrounds, illumination conditions, attitudes, angles, brightness, etc.. It can meet ReID's requirements for the dataset. Fig.~\ref{uestc-pr} shows the images of the same identity under five different CAMs.
We summarize the features in UESTC-PR different from other datasets into the following aspects:	
\begin{itemize}
	\item The size of the images is small. As shown in Fig. 5, the histogram of image height distribution indicates that the person is relatively small and fuzzy, which makes ReID difficult.
	\item The cameras are set to overlook the person. The cameras were placed at the top of the street lamp and have overlooked views. It makes simulate practical. 
	\item The datasets include a variety of moving gestures, such as walking and cycling. For example, in the process of cycling, the target is blocked by the bike. However, the difficulty of ReID increases as part of the features are lost.
	\item The datasets could be used for training the joint detection and identification task. The pedestrian picture is cropped from the original surveillance video, which can be used for the detection task, as well as the joint detection and ReID task.
\end{itemize}

At the same time, the samples in the dataset still have diversity in many aspects, such as resolution and tone. One of the five cameras has a resolution of 960*480, while the other four have a resolution of 1920*1080. Among the five cameras, one camera has a bluish tone, while the other four cameras have a normal tone, as shown in Fig. 4. A clearer statistic is shown in Table 2.

\subsection{Evaluation Protocols}
The evaluation indicators that we adopted belong to cumulative matching characteristics(CMC) and mean averaged precision(mAP). For CMC, different datasets have different calculation methods, while for the mAP, the calculation method of all datasets is same.

In this paper, we adopt 3 different evaluation indicators (i.e. the general mAP and 2 variants of the CMC on the CUHK03 dataset and the Market-1501 data set) to evaluate our model on three datasets. The detailed information about the two variants are as follows:

\textit{CUHK03}: Query and gallery sets are from different camera views. For each query, they randomly sample one instance for each gallery identity, and compute a CMC curve in the single-gallery-shot setting. The random sampling is repeated for N times and the expected CMC curve is reported.

\textit{Market-1501}: Query and gallery sets could have same camera views, but for each individual query identity, his/her gallery samples from the same camera are excluded. They do not randomly sample only one instance for each gallery identity. This means the query will always match the "easiest" positive sample in the gallery, while does not care other harder positive samples when computing.

\section{Experiments}\label{sec:5}
To evaluate the effectiveness of our approach and study the impact of various factors on person ReID, we conduct several groups of experiments on the Duke, Market-1501 and UESTC-PR dataset. Subsection \ref{sec:5.1} gives an introduction of basic experiment setting. Subsection \ref{sec:5.2} compares the effects of four different loss function (i.e. Softmax, OIM, Triplet and TOIM) on ReID. Subsection \ref{sec:5.3} conducts further researches on the strategies of selecting the feature's dimension, update rate  and negative samples. Subsection \ref{sec:5.4} compares UESTC-PR dataset with other datasets in training performance. Subsection \ref{sec:5.5} compares our approaches with existing ones.

\subsection{Experiment settings}\label{sec:5.1}
In this paper, we build the basic network by Pytorch, and use the ResNet-50 network to extract pedestrians' feature which is set 512 dimensions. The update rate $\gamma$ is 0.4 and Batchsize is 15. The length of Pooled Table is dependent on the number of IDs in training examples. For the initialization of Pooled Table, we first use Softmax loss to train on Duke dataset and then take the output of trained model to initialize Pooled Table. If some IDs are not witnessed by all the cams, the corresponding places in Pooled Table is set zero. The length of Update Table is 20. We choose the AdaDelta optimizer with an initial learning rate of 0.001. The training process contains 13 epochs.

\subsection{Effectiveness of TOIM loss}\label{sec:5.2}

In this subsection, we mainly demonstrate the effectiveness of TOIM loss through experiments, including the comparison between TOIM loss and other loss functions, the explanation of faster convergence rate of TOIM, and the improvement of accuracy after adding some tricks to TOIM.

\subsubsection{Comparison with Other Loss Functions}\label{sec:5.2.1}

Based on the settings mentioned in \ref{sec:5.1}, we evaluate the performance of 4 loss functions on the Duke, Market-1501 and UESTC-PR dataset. The result is shown in Table 3.

It's shown that TOIM gives a better performance than Softmax, OIM and Triplet loss. TOIM selects positive and negative samples by a specific strategy rather than treat all the examples equally. Therefore, when choosing a zero vector as negative sample, the training result will give a better performance of clustering by pushing all the samples away from zero points. In addition, TOIM can optimize the network without restrictions on margin due to the characteristic of OIM loss. Meanwhile, it also possesses the advantages of Triplet loss, which allows it to focus on hard samples. TOIM loss generally outperforms the other three functions.

\Figure[!htpb](topskip=0pt, botskip=0pt, midskip=0pt)[width=3.3in]{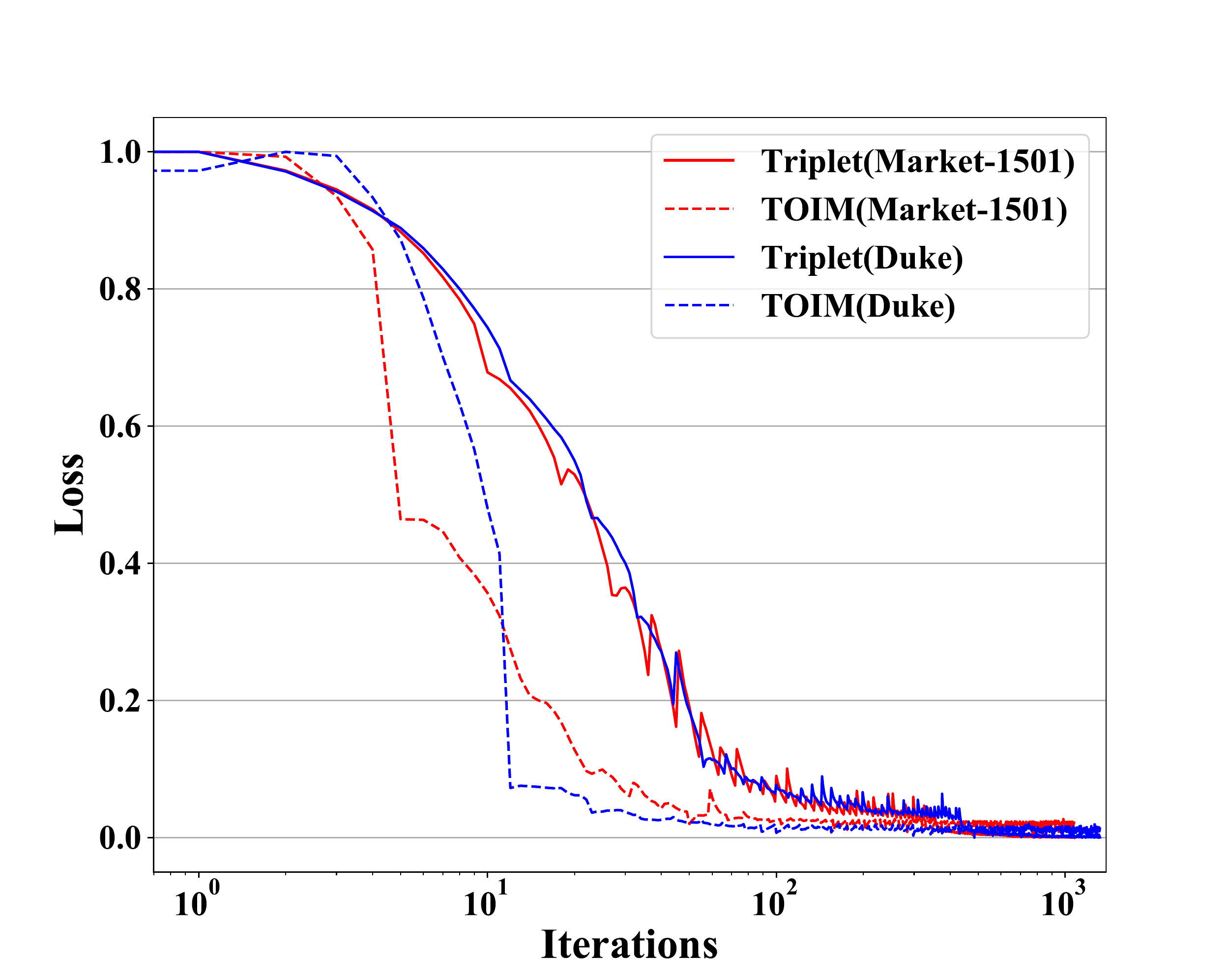}
{The convergence rates of TOIM loss and Triplet loss. The red and green line denote the Market-1501 and Duke datasets, respectively. And the solid and dasher lines represent Triplet and TOIM loss, respectively.\label{dataset_distribution}}

\begin{figure*}[!htpb]
	\centering
	\subfloat[Different values of update rate]{
		\includegraphics[width=0.3\linewidth]{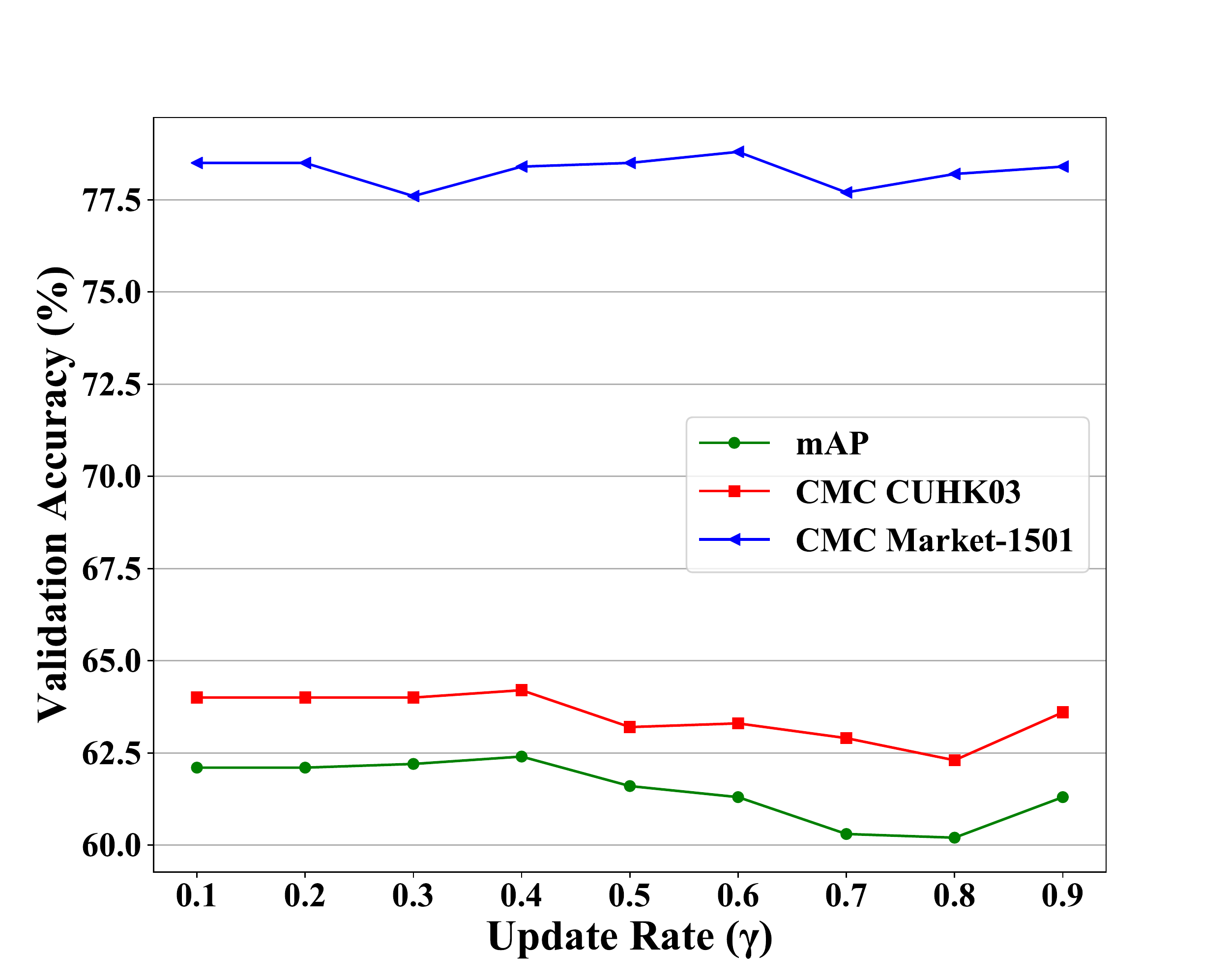}}
	\label{}\hfill
	\subfloat[Different dimensions of pedestrian feature]{
		\includegraphics[width=0.3\linewidth]{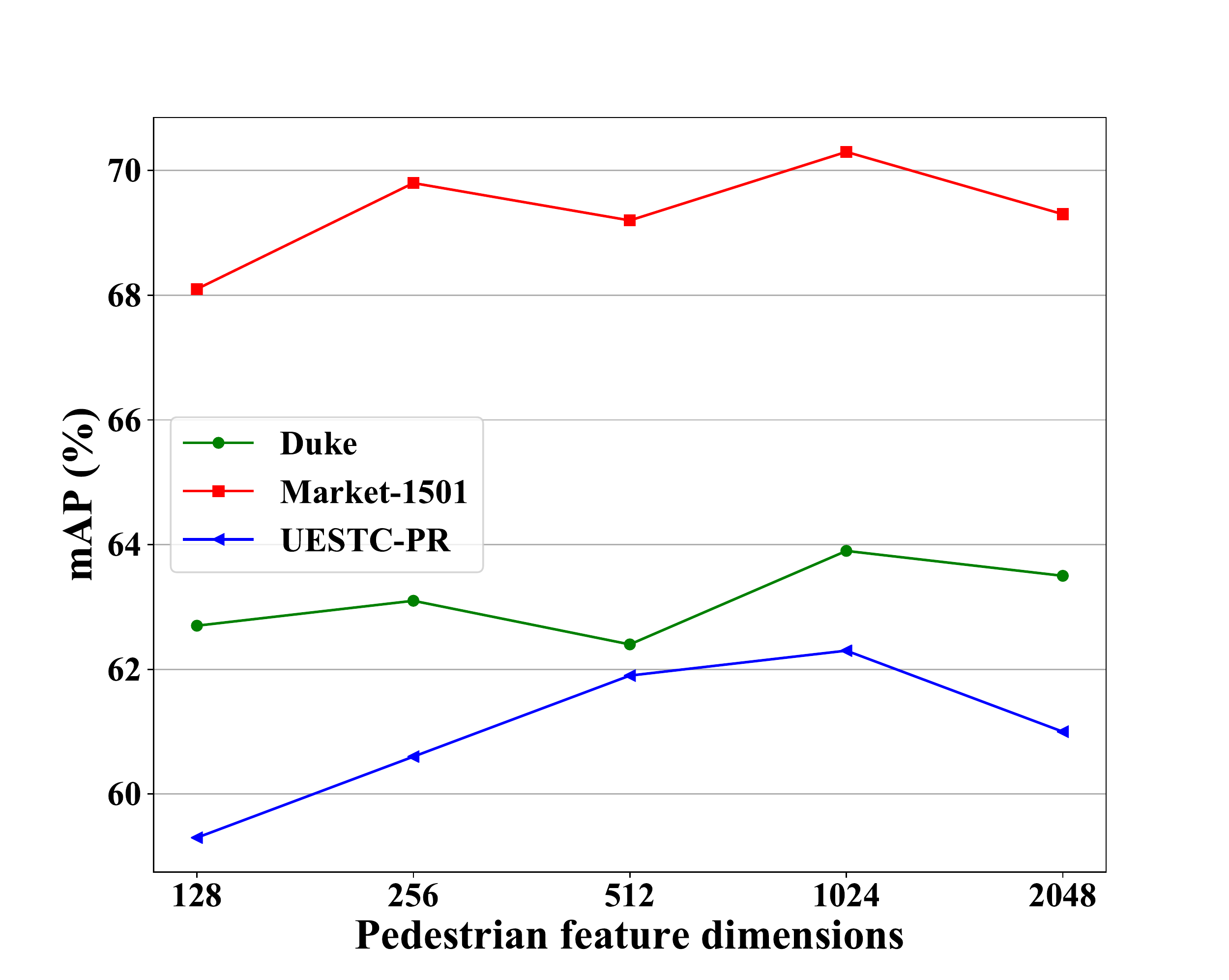}}
	\label{}\hfill
	\subfloat[Different strategies for negative samples]{
		\includegraphics[width=0.3\linewidth]{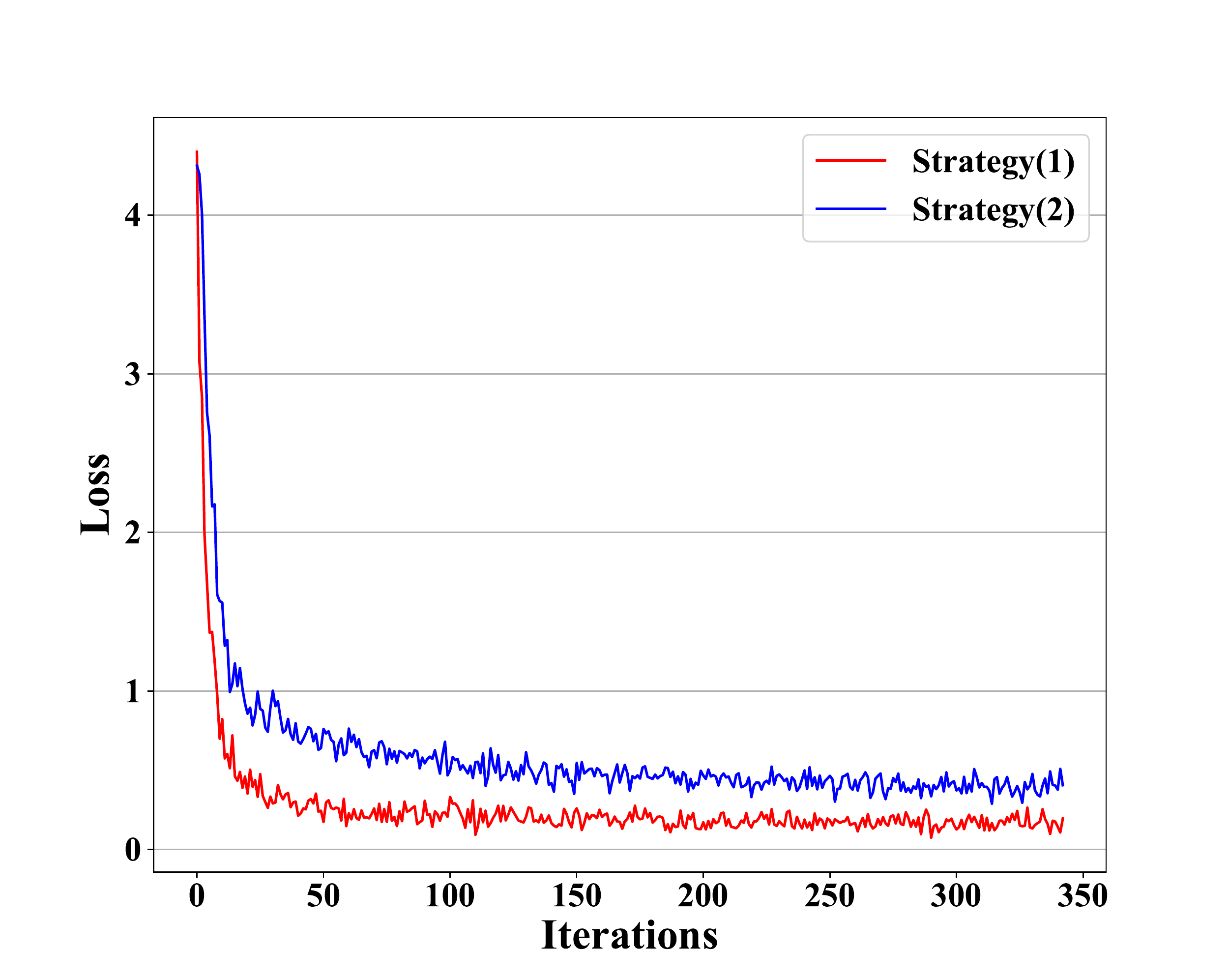}}
	\label{}\hfill
	\caption{(a),(b) Test mAP and CMC curves of different factors. (c) The loss curves of different selection strategies.}
	\label{factors}
\end{figure*}

\subsubsection{Convergence Rate}\label{5.2.2}
In this part, we would like to prove that TOIM loss converges faster than Triple loss. We choose the modified baseline in \cite{luo2019} and compare the convergence rates of TOIM and Triple loss on Market-1501 and Duke datasets, respectively. As shown in Fig. 6, the values of the ordinate are normalized, and the values of the abscissa are logarithmic. Experiment shows that the convergence rate of TOIM is faster than that of Triple loss.

As mentioned in Section \ref{sec:3.2}, in the batch construction of TOIM loss, the corresponding positive and negative sample for each anchor will be select from all person samples once the anchors are selected. However, in the batch construction of Triplet loss, we must select positive and negative samples from the current batch. Thus the batch construction determines that the TOIM loss converges faster than Triplet loss.

\subsubsection{More Tricks}\label{5.2.3}

In this part, we choose the modified baseline in \cite{luo2019} which contains six training tricks. The original loss function is formulated as:

$$L=L_{ID}+L_{Triplet}+\beta L_{C}  \eqno{(2)}  $$

To verity the effectiveness of TOIM loss function, we replace the Triplet loss function with the TOIM loss function, and other experiment settings remain unchanged. The new loss function is formulated as:

$$L_{T}=L_{ID}+L_{TOIM}+\beta L_{C}  \eqno{(3)}  $$

During the training, we initialize the Pooled Table with the output of trained model in \cite{luo2019}. After 120 training epochs, the experimental results are shown in Table 4. It can be seen that replacing the Triplet loss function with TOIM loss function can improve mAP and rank-1 accuracy on three datasets. Without re-ranking method, mAP and rank-1 can be increased by up to 1.2$\%$ and 1.5$\%$ on Duke dataset, respectively.

\begin{table}[!htbp]
	\centering
	\caption{Comparisons between our loss function and others on modified baseline\cite{luo2019}. RK stands for $k$-reciprocal re-ranking method\cite{Zhong}.}
	\setlength{\tabcolsep}{1.9mm}{
		\begin{tabular}{ccccccccc}
			\toprule
			\multicolumn{1}{c}{ \multirow{2}*{\textit{loss}} }&\multicolumn{2}{c}{Duke}& &\multicolumn{2}{c}{Market-1501}& &\multicolumn{2}{c}{UESTC-PR}\\
			\cmidrule{2-3}
			\cmidrule{5-6}
			\cmidrule{8-9}
			\multicolumn{1}{c}{}&mAP&r = 1& &mAP&r = 1& &mAP&r = 1\\
			\midrule
			$L$&76.4&86.4& &85.9&94.5& &67.3&79.1\\
			\bm{$L_{T}$}&\textbf{77.6}&\textbf{87.9}& &\textbf{87.0}&\textbf{94.4}& &\textbf{69.3}&\textbf{81.5}\\
			$L(RK)$&89.1&90.3& &94.2&95.4& &79.5&84.8\\
			\bm{$L_{T}(RK)$}&\textbf{89.2}&\textbf{90.6}& &\textbf{94.5}&\textbf{95.5}& &\textbf{84.0}&\textbf{87.2}\\
			\bottomrule
	\end{tabular}}
\end{table}

\begin{table*}[!htbp]
	\centering
	\caption{Relationship between accuracy and feature dimensions of pedestrian.}
	\setlength{\tabcolsep}{1.1mm}{
		\begin{tabular}{cccccccccccc}
			\toprule
			\multicolumn{1}{c}{ \multirow{2}*{\textit{Dimension}} }&\multicolumn{3}{c}{Duke}& &\multicolumn{3}{c}{Market-1501}& &\multicolumn{3}{c}{UESTC-PR}\\
			\cmidrule{2-4}
			\cmidrule{6-8}
			\cmidrule{10-12}
			\multicolumn{1}{c}{}&mAP&CMC CUHK03&CMC Market-1501& &mAP&CMC CUHK03&CMC Market-1501& &mAP&CMC CUHK03&CMC Market-1501\\
			\midrule
			128&62.7&64.2&78.8& &68.1&73.4&81.7& &59.3&68.4&75.4\\
			256&63.1&64.3&79.7& &69.8&75.0&83.5& &60.6&68.1&73.9\\
			512&62.4&64.2&78.4& &69.2&74.3&83.3& &61.9&69.4&73.0\\
			\textbf{1024}&\textbf{63.9}&\textbf{65.4}&\textbf{79.8}& &\textbf{70.3}&\textbf{75.2}&\textbf{84.6}& &\textbf{62.3}&\textbf{71.7}&\textbf{77.7}\\
			2048&63.5&65.1&79.9& &69.3&74.7&83.7& &61.0&70.3&78.7\\
			\bottomrule
	\end{tabular}}
\end{table*}

\subsection{Key Factors in TOIM loss}\label{sec:5.3}
During the experiment, we found that the dimension of features, update rate $\gamma$ and the selecting of negative samples have a determinant impact on the performance of our model. In this subsection, we give a detailed illustration by experiments.

\subsubsection{Update Rate}

The update rate $\gamma$ has a direct effect on the speed of updating. In this experiment, we fix the dimension to be 512 and test all the possible values of $\gamma$ on the Duke dataset (see Fig.~\ref{factors} (a)). It turns out that with the increase of $\gamma$, mAP experiences a rise before decline, reaching it peak when $\gamma$ is 0.4.

\subsubsection{Pedestrian Feature Dimensions}

The dimension of feature vectors is another key factors in result accuracy. In this experiment, $\gamma$ is 0.4 and we test the dimension as 128, 256, 512, 1024 and 2048 on the Duke, Market-1501 and UESTC-PR dataset (see Fig.~\ref{factors} (b)). The detailed result is shown in Table 5. It turns out that the accuracy reach its highest when the dimension is 1024, as a small dimension leads to a lack of information and a large dimension leads to a surplus of information.

\subsubsection{Selection Strategy of Negative Samples}

When choosing the negative samples, there are two different strategies: (1) choose samples with different identity and the smallest distance from Update Table; (2) choose samples with different identity and the smallest distance from Pooled Table. In this experiment, we set $\gamma$ as 0.4, feature dimension as 512, and compare the above 2 strategies in terms of its effects on the speed of loss function convergence. The result is shown in Fig.~\ref{factors} (c). It turns out that the former one can leads to a faster convergence.

\subsection{Effectiveness of UESTC-PR Dataset}\label{sec:5.4}

We conducted two experiments to verify the advantages of UESTC-PR dataset. We first use UESTC-PR dataset and Duke dataset to train the ResNet-50 respectively, and then use market-1501 dataset to test the network. Second, we use UESTC-PR dataset and Market-1501 dataset to train the network respectively, and then use Duke dataset to test the network. The experimental results are shown in the Fig.~\ref{transfer_learning}.  

\Figure[!htpb](topskip=0pt, botskip=0pt, midskip=0pt)[width=3.3in]{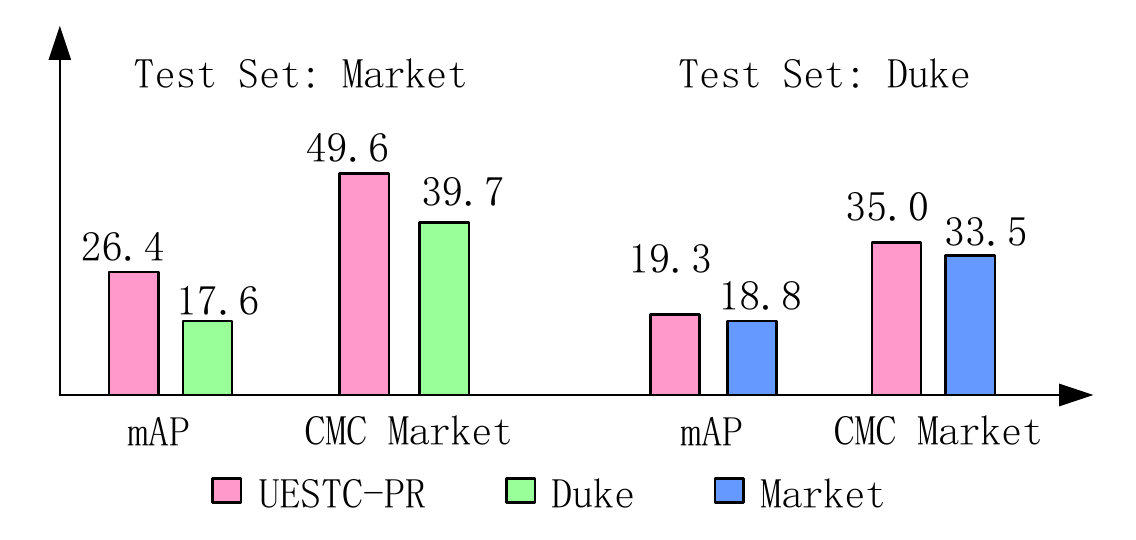}
{ MAP and CMC Market of ResNet-50 on test set of Market and Duke. The colors pink, green and blue denote the training set is UESTC-PR, Duke and Market-1501, respectively.\label{transfer_learning}}

It is obviously that ResNet-50 trained on UESTC-PR dataset works better than that trained on Market-1501 dataset and Duke dataset by more than 8.8$\%$ mAP and 0.5$\%$ mAP, respectively. It indicates that UESTC-PR dataset is more suitable for training network than Market-1501 and Duke, and the network is more robust.

\subsection{Comparison with State-of-the-art Methods}\label{sec:5.5}

In this subsection, we set $\gamma$  as 0.4, dimension of feature vector as 1024, and compare our model with the previously proposed ones on Duke dataset. The result is shown in Table 6.

It can be seen from the table that the accuracy of our proposed method only exceed a part of existing ones and still falls behind the latest methods. It is mainly on that we only improve loss function rather than the whole structure. However, our loss function outperforms others, which proves the effectiveness of TOIM. As shown in the last row of the table, the accuracy of our methods which use TOIM and add some tricks in modified baseline outperforms the state-of-the-arts methods.

\begin{table}
        \centering
         \caption{Comparison with the state-of-the-art on Duke dataset.}
         \setlength{\tabcolsep}{1.5mm}{
       \begin{tabular}{cccc}
         \toprule
         Method&mAP&CMC CUHK03&	CMC Market-1501 \\
         \midrule
         OIM \cite{Xiao2017} & - & 50.4 & 68.1 \\
         MLFN \cite{Chang2018}&	62.8&	-&	81.0\\
         PT \cite{Liu2018}&56.9&-&78.5\\
         Multi-pseudo \cite{Huang2019}&58.6&-&76.8\\
         PN-GAN \cite{Qian2018}&53.2&-&73.6\\
         Triplet\cite{Wu2019}&53.9&-&73.6\\
         \textbf{Ours}&\textbf{63.9}&\textbf{65.4}&\textbf{79.8}\\         
         SPReID \cite{Kalayeh2018}&71.0& - &84.4\\
		 Omni-directional \cite{Wu2019-Omni}&74.6& - &86.7\\
         Pyramid \cite{Zheng2018}&79.0& - &89.0\\
         BT+SB \cite{luo2019}&89.1& - &90.3\\
         \textbf{Ours(\bm{$L_{T}(RK)$})}&\textbf{89.2}& - &\textbf{90.6}\\
         \bottomrule
       \end{tabular}}
\end{table}

\section{Conclusion}\label{sec:6}
In this paper, we propose a new loss function for person ReID. It not only simplifies the construction of batch and lead to a faster convergence, but also conduct hard sample mining and improve the accuracy in ReID. As shown in experiments, it is superior to Softmax, OIM and Triplet loss. In addition we collect and annotate a large-scale benchmark dataset (UESTC-PR). Experiments show that UESTC-PR is more suitable for training person ReID model than Duke and Market-1501. As TOIM is applicable for on-line training, we plan to apply it to the training the end-to-end model for detection and re-identification. Also we would like to expand the number of pedestrians in UESTC-PR dataset to provide a larger benchmark for scholars.


\bibliographystyle{unsrt}
\nocite{*}

\bibliography{reference_access}

\begin{thebibliography}{10}

\bibitem{Zheng2016}
Liang Zheng, Yi~Yang, and Alexander~G. Hauptmann.
\newblock {Person Re-identification: Past, Present and Future}.
\newblock {\em CoRR}, abs/1610.02984(8):1--20, 2016.

\bibitem{Wu2019}
Di~Wu, Si~Jia Zheng, Xiao~Ping Zhang, Chang~An Yuan, Fei Cheng, Yang Zhao,
  Yong~Jun Lin, Zhong~Qiu Zhao, Yong~Li Jiang, and De~Shuang Huang.
\newblock {Deep learning-based methods for person re-identification: A
  comprehensive review}.
\newblock {\em Neurocomputing}, 337:354--371, 2019.

\bibitem{Cheng2016}
De~Cheng, Yihong Gong, Sanping Zhou, Jinjun Wang, and Nanning Zheng.
\newblock {Person re-identification by multi-channel parts-based CNN with
  improved triplet loss function}.
\newblock {\em CVPR}, 2016-December:1335--1344, 2016.

\bibitem{Xiao2017}
Tong Xiao, Shuang Li, Bochao Wang, Liang Lin, and Xiaogang Wang.
\newblock {Joint detection and identification feature learning for person
  search}.
\newblock {\em CVPR}, 2017-January:3376--3385, 2017.

\bibitem{He2016}
Kaiming He, Xiangyu Zhang, Shaoqing Ren, and Jian Sun.
\newblock {Deep residual learning for image recognition}.
\newblock {\em CVPR}, 2016-December:770--778, 2016.

\bibitem{Schroff2015}
Florian Schroff, Dmitry Kalenichenko, and James Philbin.
\newblock {FaceNet: A unified embedding for face recognition and clustering}.
\newblock {\em CVPR}, 07-12-June-2015:815--823, 2015.

\bibitem{Wang2016}
Faqiang Wang, Wangmeng Zuo, Liang Lin, David Zhang, and Lei Zhang.
\newblock {Joint learning of single-image and cross-image representations for
  person re-identification}.
\newblock {\em CVPR}, 2016-December:1288--1296, 2016.

\bibitem{Li2019}
Ye~Li, Guangqiang Yin, Shaoqi Hou, Jianhai Cui, and Zicheng Huang.
\newblock Spatiotemporal feature extraction for pedestrian re-identification.
\newblock pages 188--200, 2019.

\bibitem{Wu2019-Omni}
Di~Wu, Hong~Wei Yang, De~Shuang Huang, Chang~An Yuan, Xiao Qin, Yang Zhao,
  Xin~Yong Zhao, and Jian~Hong Sun.
\newblock {Omnidirectional feature learning for person re-identification}.
\newblock {\em IEEE Access}, 7:28402--28411, 2019.

\bibitem{Zheng2017}
Zhedong Zheng, Liang Zheng, and Yi~Yang.
\newblock {Unlabeled Samples Generated by GAN Improve the Person
  Re-identification Baseline in Vitro}.
\newblock {\em ICCV}, 2017-October:3774--3782, 2017.

\bibitem{Zheng2015}
Liang Zheng, Liyue Shen, Lu~Tian, Shengjin Wang, Jingdong Wang, and Qi~Tian.
\newblock {Scalable Person Re-identification : A Benchmark University of Texas
  at San Antonio}.
\newblock {\em ICCV}, pages 1116--1124, 2015.

\bibitem{luo2019}
Hao Luo, Youzhi Gu, Xingyu Liao, Shenqi Lai, and Wei Jiang.
\newblock {Bag of Tricks and A Strong Baseline for Deep Person
  Re-identification}.
\newblock 2019.

\bibitem{Zhong}
Zhun Zhong, Liang Zheng, Donglin Cao, and Shaozi Li.
\newblock {Re-ranking Person Re-identification with k-reciprocal Encoding}.
\newblock pages 1318--1327.

\bibitem{Chang2018}
Xiaobin Chang, Timothy~M. Hospedales, and Tao Xiang.
\newblock {Multi-level Factorisation Net for Person Re-identification}.
\newblock {\em Proceedings of the IEEE Computer Society Conference on Computer
  Vision and Pattern Recognition}, pages 2109--2118, 2018.

\bibitem{Liu2018}
J.~{Liu}, B.~{Ni}, Y.~{Yan}, P.~{Zhou}, S.~{Cheng}, and J.~{Hu}.
\newblock Pose transferrable person re-identification.
\newblock pages 4099--4108, June 2018.

\bibitem{Huang2019}
Yan Huang, Jingsong Xu, Qiang Wu, Zhedong Zheng, Zhaoxiang Zhang, and Jian
  Zhang.
\newblock {Multi-pseudo regularized label for generated data in person
  re-identification}.
\newblock {\em IEEE Transactions on Image Processing}, 28:1391--1403, 2019.

\bibitem{Qian2018}
Xuelin Qian, Yanwei Fu, Tao Xiang, Wenxuan Wang, Jie Qiu, Yang Wu, Yu~Gang
  Jiang, and Xiangyang Xue.
\newblock {Pose-normalized image generation for person re-identification}.
\newblock {\em Lecture Notes in Computer Science (including subseries Lecture
  Notes in Artificial Intelligence and Lecture Notes in Bioinformatics)}, 11213
  LNCS:661--678, 2018.

\bibitem{Kalayeh2018}
Mahdi~M. Kalayeh, Emrah Basaran, Muhittin Gokmen, Mustafa~E. Kamasak, and
  Mubarak Shah.
\newblock {Human Semantic Parsing for Person Re-identification}.
\newblock {\em Proceedings of the IEEE Computer Society Conference on Computer
  Vision and Pattern Recognition}, pages 1062--1071, 2018.

\bibitem{Zheng2018}
Feng Zheng, Cheng Deng, Xing Sun, Xinyang Jiang, Xiaowei Guo, Zongqiao Yu,
  Feiyue Huang, and Rongrong Ji.
\newblock {Pyramidal Person Re-IDentification via Multi-Loss Dynamic Training}.
\newblock 2018.

\bibitem{DBLP:journals/corr/ZhengYH16}
Liang Zheng, Yi~Yang, and Alexander~G. Hauptmann.
\newblock Person re-identification: Past, present and future.
\newblock {\em CoRR}, abs/1610.02984, 2016.

\bibitem{Tian2015}
Yonglong Tian, Ping Luo, Xiaogang Wang, and Xiaoou Tang.
\newblock {Pedestrian detection aided by deep learning semantic tasks}.
\newblock In {\em CVPR}, 2015.

\bibitem{Zeiler2012}
Matthew~D. Zeiler.
\newblock {ADADELTA: An Adaptive Learning Rate Method}.
\newblock 2012.

\bibitem{Liu2017}
Hao Liu, Jiashi Feng, Zequn Jie, Karlekar Jayashree, Bo~Zhao, Meibin Qi,
  Jianguo Jiang, and Shuicheng Yan.
\newblock {Neural Person Search Machines}.
\newblock {\em Proceedings of the IEEE International Conference on Computer
  Vision}, 2017-October:493--501, 2017.

\bibitem{Bolle2005}
{The relation between the ROC curve and the CMC}.
\newblock {\em Proceedings - Fourth IEEE Workshop on Automatic Identification
  Advanced Technologies, AUTO ID 2005}, 2005:15--20, 2005.

\bibitem{Munjal2019}
Bharti Munjal, Sikandar Amin, Federico Tombari, and Fabio Galasso.
\newblock {Query-guided End-to-End Person Search}.
\newblock pages 811--820, 2019.

\bibitem{Chen2018}
Di~Chen, Shanshan Zhang, Wanli Ouyang, Jian Yang, and Ying Tai.
\newblock {Person search via a mask-guided two-stream CNN model}.
\newblock {\em Lecture Notes in Computer Science (including subseries Lecture
  Notes in Artificial Intelligence and Lecture Notes in Bioinformatics)}, 11211
  LNCS:764--781, 2018.

\bibitem{Li2014}
Wei Li, Rui Zhao, Tong Xiao, and Xiaogang Wang.
\newblock {DeepReID: Deep filter pairing neural network for person
  re-identification}.
\newblock {\em CVPR}, pages 152--159, 2014.

\bibitem{Zheng2009}
Wei~Shi Zheng, Shaogang Gong, and Tao Xiang.
\newblock {Associating groups of people}.
\newblock {\em British Machine Vision Conference, BMVC 2009 - Proceedings},
  (1), 2009.

\bibitem{Hirzer2011}
Martin Hirzer, Csaba Beleznai, Peter~M. Roth, and Horst Bischof.
\newblock {Person re-identification by descriptive and discriminative
  classification}.
\newblock {\em Lecture Notes in Computer Science (including subseries Lecture
  Notes in Artificial Intelligence and Lecture Notes in Bioinformatics)}, 6688
  LNCS:91--102, 2011.

\bibitem{Li2013}
Wei Li and Xiaogang Wang.
\newblock {Locally aligned feature transforms across views}.
\newblock {\em CVPR}, pages 3594--3601, 2013.

\bibitem{Gray2007}
Doug Gray, Shane Brennan, and Hai Tao.
\newblock {Evaluating appearance models for recognition, reacquisition, and
  tracking}.
\newblock {\em 10th International Workshop on Performance Evaluation for
  Tracking and Surveillance (PETS),}, 3:41--47, 2007.

\bibitem{Enhancement}
Feature Enhancement.
\newblock {Supplementary Material Diversity Regularized Spatiotemporal
  Attention for Video-based Person Re-identification}.
\newblock (3):1--2.

\bibitem{Tran2015}
Du~Tran, Lubomir Bourdev, Rob Fergus, Lorenzo Torresani, and Manohar Paluri.
\newblock {Learning spatiotemporal features with 3D convolutional networks}.
\newblock {\em ICCV}, 2015 International Conference on Computer Vision, ICCV
  2015:4489--4497, 2015.

\bibitem{Liu2015}
Kan Liu, Bingpeng Ma, Wei Zhang, and Rui Huang.
\newblock {A spatio-temporal appearance representation for viceo-based
  pedestrian re-identification}.
\newblock {\em ICCV}, 2015 International Conference on Computer Vision, ICCV
  2015:3810--3818, 2015.

\bibitem{Xiao2019}
Jimin Xiao, Yanchun Xie, Tammam Tillo, Kaizhu Huang, Yunchao Wei, and Jiashi
  Feng.
\newblock {IAN: The Individual Aggregation Network for Person Search}.
\newblock {\em Pattern Recognition}, 87:332--340, 2019.

\bibitem{Zhang2015}
Ning Zhang, Manohar Paluri, Yaniv Taigman, Rob Fergus, and Lubomir Bourdev.
\newblock {Beyond frontal faces: Improving Person Recognition using multiple
  cues}.
\newblock {\em CVPR}, 07-12-June-2015:4804--4813, 2015.

\bibitem{Zhang2016}
Li~Zhang, Tao Xiang, and Shaogang Gong.
\newblock {Learning a discriminative null space for person re-identification}.
\newblock {\em CVPR}, 2016-December:1239--1248, 2016.

\end{thebibliography}

\begin{IEEEbiography}[{\includegraphics[width=1in,height=1.25in,clip,keepaspectratio]{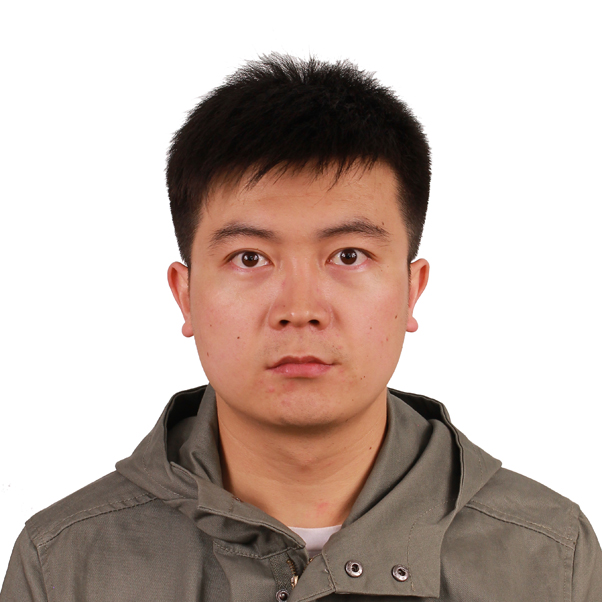}}]{Ye Li} received a bachelor's degree from the School of Information Science and Technology of Hainan University in 2014. In 2014, He went to Taiwan ilan university as an exchange student.He is currently pursuing the Ph.D.
degree in Department of Electronic and communication engineering,
University of Electronic Science and Technology of China,
Chengdu,Sichuan

\end{IEEEbiography}

\begin{IEEEbiography}[{\includegraphics[width=1in,height=1.25in,clip,keepaspectratio]{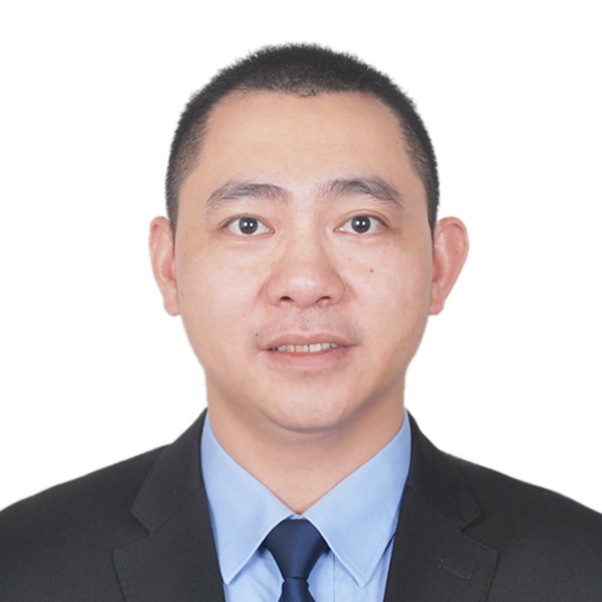}}]{Guangqiang Yin} is a professor at the University of Electronic Science and Technology of China (UESTC). His research interests include computer vision related artificial intelligence techniques and applications, and computer modeling of properties of condensed matter.
\end{IEEEbiography}

\begin{IEEEbiography}[{\includegraphics[width=1in,height=1.25in,clip,keepaspectratio]{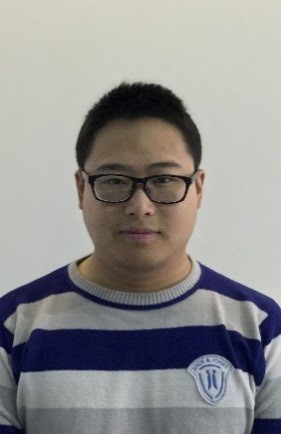}}]{Chunhui Liu} received his bachelor's degree in engineering from the School of Information Engineering, Southwest University of Science and Technology in 2017. He is currently pursuing a master's degree in Information and Communication Engineering at University of Electronic Science and Technology of China. He has participated in the collection and processing of human body electromechanical signals, high frequency wireless signal transmission, embedded development and other research. Now his research interests include image and video processing, computer vision, operating systems, biomedical engineering.

\end{IEEEbiography}

\begin{IEEEbiography}[{\includegraphics[width=1in,height=1.25in,clip,keepaspectratio]{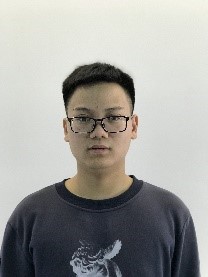}}]{Xiaoyu Yang} received a bachelor's degree from the School of Information Science and Technology of Hainan University in 2014. His previous work focused on embedded development research. He is currently involved in the study of pedestrian multi-attribute recognition and pedestrian recognition. His research interests include image processing, computer vision, deep learning, target detection.

\end{IEEEbiography}

\begin{IEEEbiography}[{\includegraphics[width=1in,height=1.25in,clip,keepaspectratio]{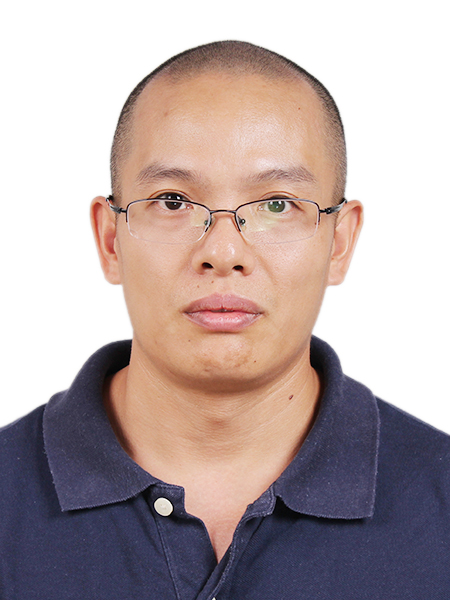}}]{Zhiguo Wang} received his Ph.D.  degree in materials physics and chemistry from UESTC in 2008. He is a professor at the University of Electronic Science and Technology of China (UESTC). He graduated from Sichuan University in China and received M.S. in condensed matter physics in 2002. He got. His research interests include computer vision related artificial intelligence techniques and applications, and computer modeling of properties of condensed matter.

\end{IEEEbiography}

\EOD

\end{document}